\documentclass[journal]{IEEEtran}
\usepackage[noend]{algpseudocode}
\usepackage{algorithmicx,algorithm}
\usepackage{bm}
\usepackage{graphicx}
\usepackage{subfigure}
\usepackage{amssymb}
\newcommand{\tabincell}[2]{\begin{tabular}{@{}#1@{}}#2\end{tabular}}
\usepackage[colorlinks, linkcolor=black, anchorcolor=black, citecolor=black]{hyperref}

\ifCLASSINFOpdf
\else
\fi
\begin{document}
%
\title{Highlight Every Step: Knowledge Distillation \\ via Collaborative Teaching}
%
%
%

\author{Haoran~Zhao, Xin~Sun, Junyu~Dong, Changrui~Chen and Zihe~Dong
       
\thanks{This work was supported in part by National Natural Science Foundation of China under Project No.U1706218 and 41576011. } 

\thanks{H Zhao, X Sun, J Dong, C Chen and Z Dong are with the Department of Computer Science and Technology, Ocean University of China, Qingdao, Shandong Province, 266100 China (e-mail:zhaohaoran@stu.ouc.edu.cn; sunxin@ouc.edu.cn; dongjunyun@ouc.edu.cn; ccr@stu.ouc.edu.cn; dongzihe@stu.ouc.edu.cn)}

\thanks{Manuscript received April 19, 2005; revised August 26, 2015.}}

%
%

\markboth{Journal of \LaTeX\ Class Files,~Vol.~14, No.~8, August~2015}%
{Shell \MakeLowercase{\textit{et al.}}: Bare Demo of IEEEtran.cls for IEEE Journals}
%



\maketitle

\begin{abstract}
High storage and computational costs obstruct deep neural networks to be deployed on resource-constrained devices. Knowledge distillation aims to train a compact student network by transferring knowledge from a larger pre-trained teacher model. However, most existing methods on knowledge distillation ignore the valuable information among training process associated with training results. In this paper, we provide a new Collaborative Teaching Knowledge Distillation (CTKD) strategy which employs two special teachers. Specifically, one teacher trained from scratch (i.e., scratch teacher) assists the student step by step using its temporary outputs. It forces the student to approach the optimal path towards the final logits with high accuracy. The other pre-trained teacher (i.e., expert teacher) guides the student to focus on a critical region which is more useful for the task. The combination of the knowledge from two special teachers can significantly improve the performance of the student network in knowledge distillation. The results of experiments on CIFAR-10, CIFAR-100, SVHN and Tiny ImageNet datasets verify that the proposed knowledge distillation method is efficient and achieves state-of-the-art performance.
\end{abstract}

\begin{IEEEkeywords}
Neural Networks Compression, Knowledge Distillation, Computer Vision, Deep Learning.
\end{IEEEkeywords}

%
\IEEEpeerreviewmaketitle

\section{Introduction}
%
%
%
%
\IEEEPARstart
{R}{ecently,} deep neural networks achieved superior performance in a variety of applications such as computer vision~\cite{He_2016_CVPR}\cite{Szegedy_2015_CVPR}\cite{8723079}\cite{6680765} and natural language processing~\cite{Antol_2015_ICCV}\cite{Noh_2016_CVPR}. However, along with high-performance, the deep neural network's architecture becomes much deeper and wider which requires a high cost of computation and memory in inference. It is a great burden to deploy these models on edge-computing systems such as embedded devices and mobile-phones. Therefore, many methods~\cite{Song2015Deep}\cite{Hassibi1993Second}\cite{Jaderberg2014Speeding}\cite{Cun1989Optimal}\cite{Lin2019Towards} are proposed to reduce the deep neural network's computational complexity and high  storage. Some lightweight networks like Inception~\cite{Szegedy2016Inception}, MobileNet~\cite{Howard2017MobileNets}, ShuffleNet~\cite{Zhang2017ShuffleNet}, SqueezeNet~\cite{DBLP:journals/corr/IandolaMAHDK16} and Condense-Net~\cite{DBLP:conf/cvpr/HuangLMW18} have been proposed to reduce the network size as much as possible under the condition of keeping a high recognition accuracy. All the above mentioned methods focus on physically reducing internal redundancy of the model to obtain a shallow and thin architecture. Nevertheless, how to train the reduced network with high performance is yet an unresolved issue.

\begin{figure}
  \centering
  \includegraphics[width=7.5cm]{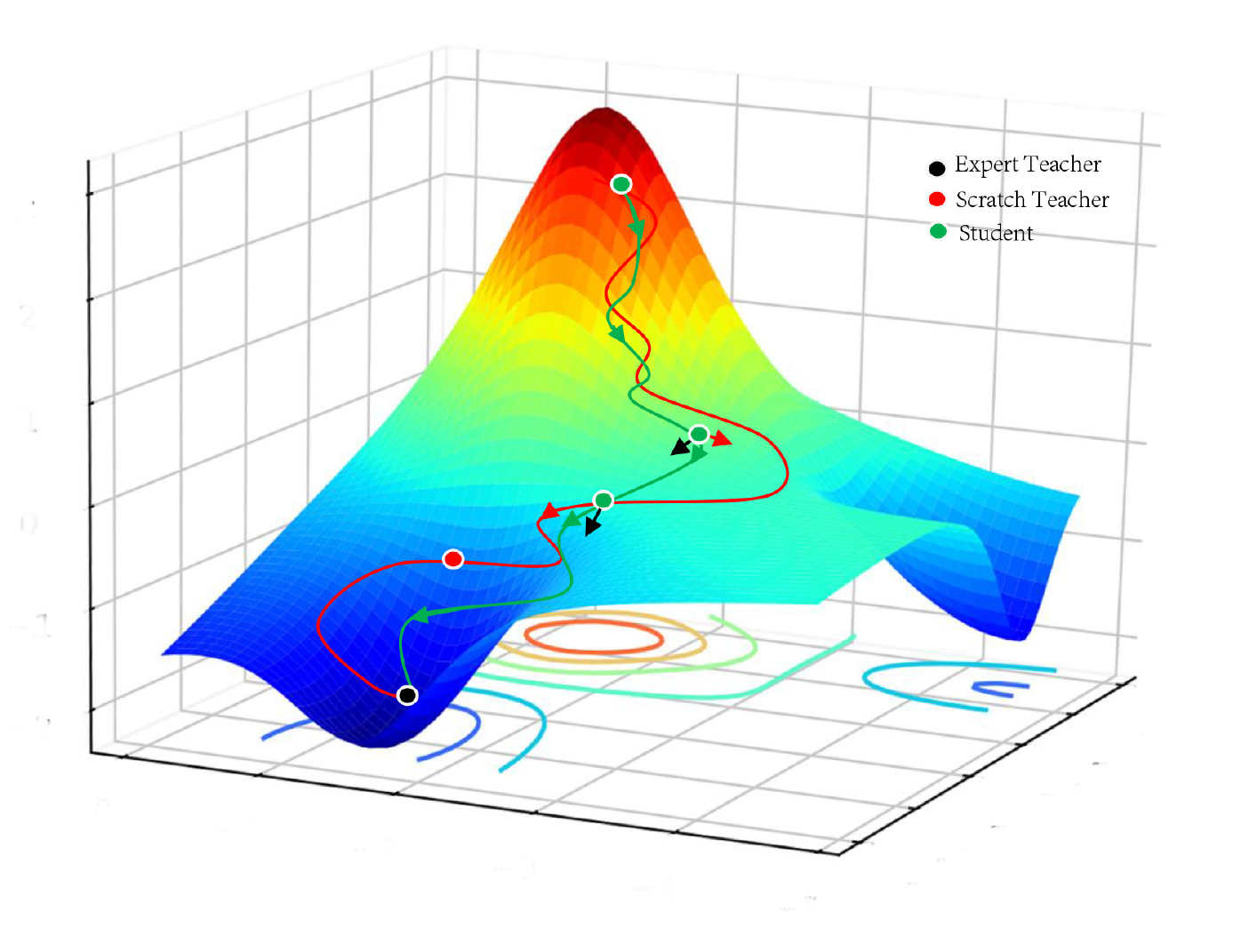}\\
  \caption{Illustration of our collaborative teaching knowledge distillation (CTKD) strategy. We illustrate the optimization process of student network (green ball) under the collaborative guidance of scratch teacher (red ball) and expert teacher (black ball). The red and green line represent the optimization path of scratch teacher and student network. And the expert teacher has already reached the local optimum. The student network starts the optimization process with scratch teacher and expert teacher.
  }\label{fig:structure}
\end{figure}

It is therefore critical to effectively train a compact neural network, and this research issue attracts more and more attention~\cite{DBLP:journals/corr/abs-1811-05072}\cite{DBLP:journals/corr/HanPTD15}, of which knowledge distillation is considered to be able to provide a practical way. Generally speaking, the distilling technique using the teacher-student strategy commonly trains a compact and shallow student network under the guidance of a complicated large teacher network. It is an effective approach to produce a compact neural network with performance close to the complicated teacher network. Once trained, this compact neural network can be directly deployed on resource-constrained devices. Knowledge Distillation (KD)~\cite{Hinton2015Distilling} uses a pre-trained teacher's soften outputs as dark knowledge to supervise the training process of student network. It assumes the knowledge as a learned mapping from inputs to outputs, and transfers the knowledge by training the student with the teacher's outputs as targets. The hint-based training approach~\cite{Romero2015FitNets} and attention transfer~\cite{Zagoruyko2016Paying} are devised to transfer the knowledge of intermediate layers from the teacher network to student network. Moreover, these approaches based on the teacher-student strategy can be combined with any physical methods. For example, network quantization can be combined with knowledge distillation~\cite{Mishra2017Apprentice} to obtain a low-precision student network with high performance. Despite the very promising results, current methods only utilize different forms of knowledge limited in the pre-trained teacher network which may ignore the valuable knowledge in the training process of the teacher network.

\begin{figure*}
  \centering
  \includegraphics[width=16cm]{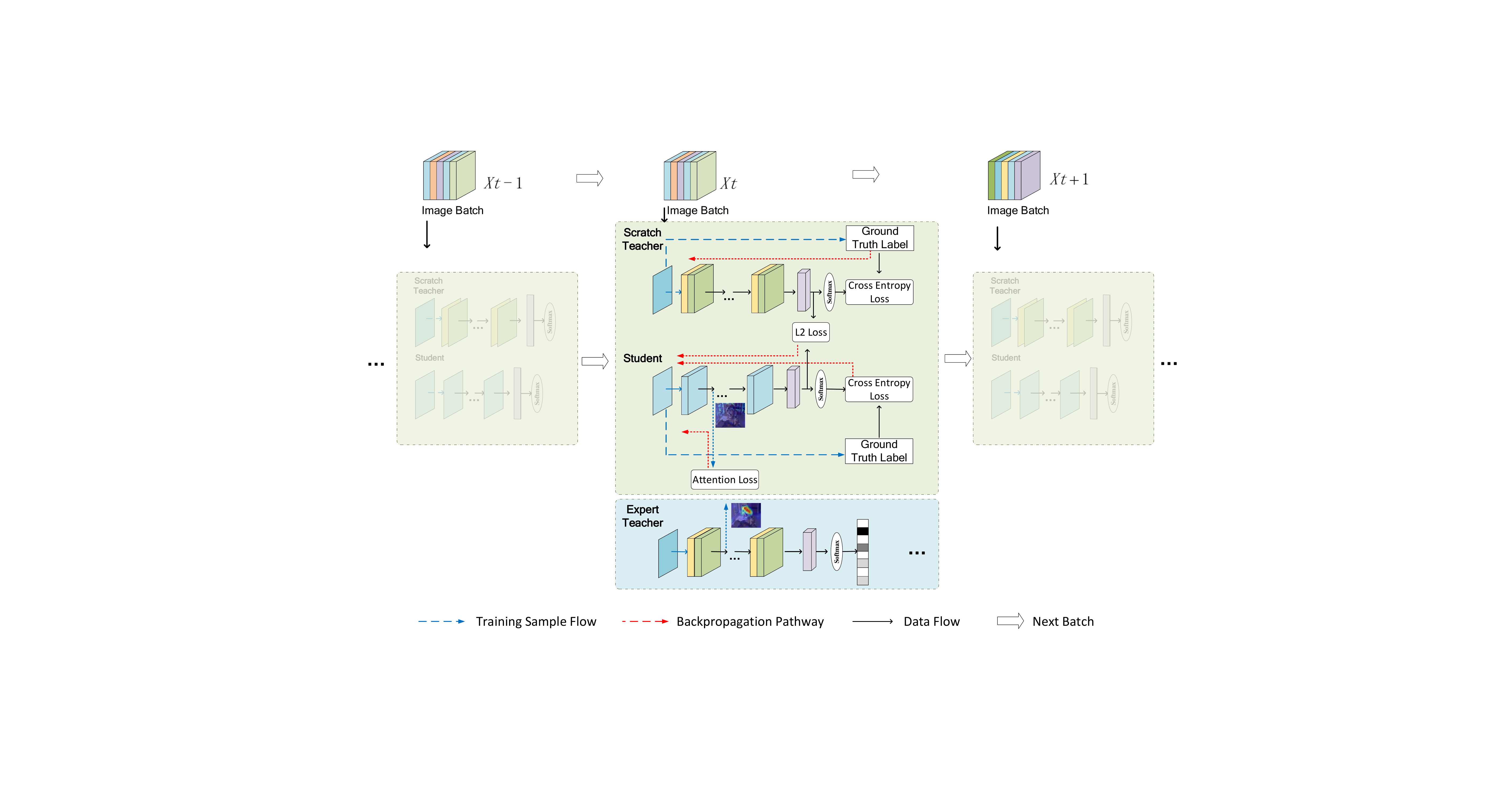}\\
  \caption{Illustration of the architecture. The scratch teacher collaboratively trains with the student network from scratch. We use standard cross-entropy loss for scratch teacher network and student network to learn the ground truth respectively. Moreover, the distillation loss supervises the training of student network by every step. The expert teacher (pre-trained) guides the student network to focus on critical region through intermediate-level attention maps.}\label{fig:optimization}
\end{figure*}

In this paper, we optimize the student network with the distilled knowledge from both a scratch teacher and an expert teacher. As illustrated in Figure \ref{fig:structure}, the expert teacher (black ball) has already reached the local optimum and the scratch teacher (red ball) continuously trains with the student (green ball) from scratch. In the process of optimization, the scratch teacher pulls the student towards its optimal path (red arrow), and the expert teacher guides the student to focus on the key region which is more useful for the tasks (black arrow). In such collaborative teaching, the student reaches the local optimum with performance close to the teachers. Our motivation is that the scratch teacher and expert teacher can provide different supervisory information which can be fully utilized through collaborative training. Namely, we use the scratch teacher to jointly train with the student network in the whole training process. Due to the strong ability of the scratch teacher, it can guide the student towards the final logits with high accuracy  step by step along the optimization path. However, the scratch teacher also wastes a large number of steps to optimize the path where the expert teacher has gone. This is the reason that we use the additional expert teacher to provide intermediate-level hints for the training of the student network. As shown in Figure \ref{fig:optimization}, the scratch teacher provides temporary logits to supervise the whole training process of the student in the pale green rectangular frame. Meanwhile, the pre-trained teacher provides attention maps from the middle of DNNs to constrain lower layers of the student. In such manner, the compact student network can produce  performance close to the teacher.

We verify our proposed Collaborative Teaching Knowledge Distillation (CTKD) method on CIFAR-10, CIFAR-100, SVHN and Tiny ImageNet datasets. The experimental results show that our method effectively improves the student's performance in knowledge distillation. Our contributions in this paper are summarized as follows:

\begin{itemize}
\item We propose a novel teacher-student knowledge distillation strategy using two teachers; it combines both the path knowledge towards the final logits with high accuracy and the intermediate-level attention knowledge for lower layers. In addition to the final outputs from the pre-trained teacher, the proposed architecture can continuously supervise the student network. 
\item We analyze the importance of both knowledge from the two teachers, and we investigate the effect of attention maps distilled from a deep teacher network on the small student network.
\item We verify our method on several public datasets. Experiments show that our method can significantly improve the performance of student networks in knowledge distillation. 
\end{itemize}

The rest of this paper is organized as follows. Related work is reviewed in Part II. And we present the proposed knowledge distillation architecture using two teachers in Part III. Experimental results are presented in Part IV. Finally, Part V concludes this paper.

\section{Related Work}
Deep neural networks have demonstrated extraordinary performance on various computer vision and machine learning tasks~\cite{6324460}\cite{7875137}\cite{7776867}. Traditional handcrafted features~\cite{perronnin10eccv} for computer vision tasks are replaced by deep neural networks which have strong ability at fitting the complicated feature-space distributions. Recently, deep neural networks become predominant in the large-scale competitions~\cite{russakovsky2015imagenet}\cite{7299291}\cite{6844850}. Researchers design much deeper and wider networks~\cite{simonyan2014very}\cite{he2016deep}\cite{huang2017densely} to further improve classification accuracy, and also tend to discover network architectures automatically~\cite{xie2017genetic}\cite{zoph2016neural}\cite{liu2018progressive}. Powered by the powerful computational resources of the work stations and GPU clusters, it is possible to train and deploy such complicated deep networks. However, the resource-constrained  devices are almost impossible to launch such complicated CNNs due to the computational complexity and high storage. For instance, over 232MB of memory and over $7.24 \times 10^{8}$  multiplications are demanded for processing one image using AlexNet~\cite{Krizhevsky2012ImageNetCW}, which cannot be tolerated by these devices~\cite{Wang2018AdversarialLO}. Therefore, compact deep models with similar accuracies are urgently expected.

Indeed, training phase of the deep neural networks is usually performed on CPU and/or GPU clusters. The challenging we really need to face is the deployment of trained models on inference systems such as resource constrained devices. During the past few years, many researchers have been studying how to deploy these deep neural networks in practice~\cite{DBLP:journals/corr/abs-1811-05072}\cite{7464340}\cite{7332782}. 
The number of parameters usually represents the model complexity, but not all parameters contribute to the performance in inference stage~\cite{Urban2016Do}\cite{DBLP:journals/tmm/DongHMKWZCI18}\cite{DBLP:conf/ijcai/LuYF17}. Model compression
techniques \cite{DBLP:conf/icml/WeinbergerDLSA09}\cite{Denil2013Predicting}\cite{DBLP:journals/corr/ChenWTWC15}\cite{DBLP:journals/corr/ZhangZMH014} have emerged to obtain a small model which retains the accuracy of a large one. In the following, we will briefly describe the most related works on network model compression and acceleration.

\textbf{DNNs compression and acceleration} are important to the real-time applications which has gained increasing interests. These methods can be roughly divided into parameter pruning,  low-rank decomposition and knowledge distillation. Parameter pruning~\cite{Cun1989Optimal}\cite{Hao2016Pruning}\cite{DBLP:journals/corr/abs-1901-07827} removes redundant weights from the pre-trained network model, which can keep the accuracy of the larger model if the prune ratio is set properly. Recently, channel pruning, which is better compatibility with off-the-shelf computing libraries, has become increasingly popular. Luo et al.~\cite{luo2017thinet} propose to use the statistics of next layer to select the channel to be pruned. However, parameter pruning approaches require many iterations to converge and we also need to manually set the pruning threshold. Low-rank decomposition~\cite{Denil2013Predicting}\cite{Kim2015Compression}\cite{DBLP:conf/nips/RenZ0PCZL018} decomposes the original convolution kernel in DNNs model by using matrix decomposition technique. But such kind of methods increase the layers of the model, and are easy to cause the vanishing gradient during the training process. Both parameter pruning and low-rank decomposition usually lead to large accuracy drops, thus fine-tuning is required to alleviate those drops~\cite{liu2017learning}\cite{8320372}. 

Besides, the reinforcement learning algorithm can be used for designing networks such as Neural Architecture Search \cite{Bello2016Neural} and MetaQNN \cite{Baker2016Designing}. The network itself could search the efficient structure without manually setting. However, these models only focus on high performance rather than the size of model.

\textbf{Knowledge distillation methods} are used to reduce the computational cost in test stage. These approaches usually utilize the teacher-student strategy, where a large pre-trained teacher network supervises the training of a small student network, for facilitating the deployment at test time. Bucilua et al.~\cite{Bucilua:2006:MC:1150402.1150464} pioneer these series of methods in model compression. They attempt to transfer the knowledge from an ensemble of heterogeneous models to a small model.  Ba et al.~\cite{Caruana2013Do} extend this method through forcing the wider and shallower student network to mimic the teacher network's logits before the softmax. Hinton et al.~\cite{Hinton2015Distilling} firstly provide the concept of knowledge distillation by introducing a hyper-parameter temperature to divide the logits before softmax. The student network is forced to imitate the distribution of teacher network's soft targets which contains more information than one-hot targets. In other words, the student's fitting goal is no longer the one-hot vector (ground-truth) which is too strict, but learns towards the teacher's soften vector which most often with correct prediction. Besides that, researchers attempt to get more supervised information from teacher network. Romero et al.~\cite{Romero2015FitNets} introduces a new metric of intermediate features between teacher and student networks. Zagoruyko et al.~\cite{Zagoruyko2016Paying} uses attention features from intermediate layers as the supervised information. Yim et al.~\cite{Yim2017A} proposes a new method using gram matrix to fit the relationship between layers and students imitate the process of solving problems by teachers. Polino et al.~\cite{polino2018model} and Mishra et al.~\cite{Mishra2017Apprentice} reduce bit precision of weights and activations by combining KD and network quantization. Xu et al.~\cite{xu2017training} use a conditional adversarial network to learn the loss function for KD. A noise-based regularizer has been proposed for KD in ~\cite{sau2016deep} and Lopes et al.~\cite{lopes2017data} use the teacher model to provide metadata for data-free KD.

Recently, researchers note that it is effective in improving a teacher model itself by self-distillation~\cite{Bagherinezhad2018Label}~\cite{Furlanello2018Born}, namely, a few models with the same architecture are trained one by one. The deep networks can be optimized in many generations, in which the next model is under the supervision of the previous one. Moreover, knowledge distillation also has been applied to other applications , such as object detection~\cite{Li_2017_CVPR}, pedestrian re-identification~\cite{chen2018darkrank}, semantic segmentation~\cite{xie2018improving}. There also exists works that unify KD with privileged information~\cite{pechyony2010theory}\cite{vapnik2015learning}\cite{vapnik2009new} as generalized distillation where a teacher is pre-trained by taking as input privileged information.

There are also some theoretical and systematic studies about how and why knowledge distillation improves neural network training. Furlanello et al.~\cite{Furlanello2018Born} analyze the success of knowledge distillation through gradients on the soft-target part which acts as sampling weight based on the teacher's confidence in its maximum value. Zhang et al.~\cite{zhang2018deep} investigate knowledge distillation via the posterior
entropy and prove that soft-targets is a much more informed choice than blind entropy regularization. 

All the above methods use only one single teacher to provide supervised information. Recently, Shan et al.~\cite{Shan2017Learning} attempt to combine the knowledge of multiple teacher networks in the intermediate representations. And Shen et al.~\cite{DBLP:journals/corr/abs-1811-02796} aim at learning a compact student model capable of handing the 'super' task from multiple teachers. Mishra et al. \cite{Mishra2017Apprentice} propose a new perspective view to combine network quantization with knowledge distillation. They jointly train a teacher network and a student from scratch using knowledge distillation. Zhou et al.~\cite{zhou2017Rocket} also provide a similar scheme where the student network and the teacher network share the lower layers and train simultaneously. The previous study~\cite{zhou2017Rocket} differs from ours in that their one-stage method sharing lower layers between teacher and student network and without using additional guidance from pre-trained teacher network, while our two-stage architecture combine intermediate-level features from teacher network with training process from teacher. It means that both the path knowledge towards the final logits with high accuracy and the intermediate-level attention knowledge for lower layers are used in the training process.

\section{Method}
The core idea of our method is to jointly train the student network using two teachers, i.e., one expert teacher trained in advance provides attention maps as the intermediate-level supervised information, the other scratch teacher with random initialization provides optimal path knowledge which towards final logits with high accuracy.

\subsection{Motivation}
Existing knowledge distillation methods \cite{Hinton2015Distilling} let the student network simply mimic the final outputs of the teacher network. However, in the case of the DNNs, there are many ways to generate the final outputs. So the student network might go around and close to the final targets in various ways. In this sense, mimicking the outputs of the teacher network can be a hard constraint for the student network. We propose the Collaborative Teaching Knowledge Distillation (CTKD) method to remedy such situation.

\begin{figure}
  \centering
  \includegraphics[width=8.5cm]{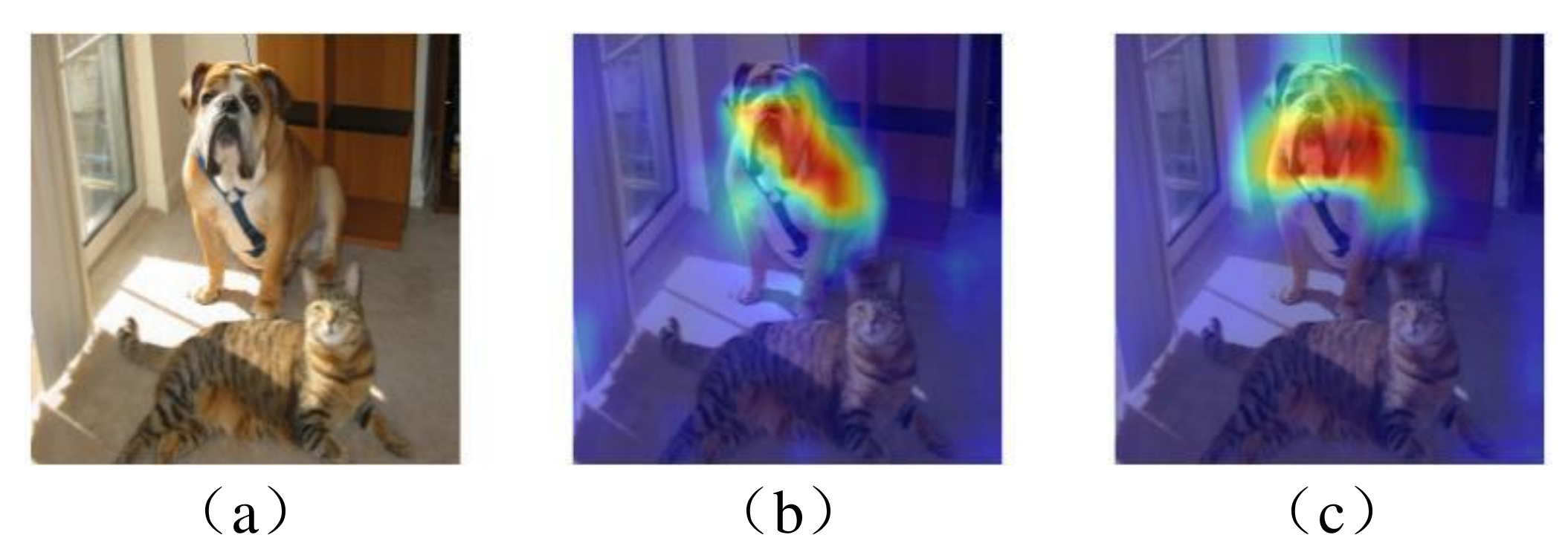}\\
  \caption{Visualization of top activation attention maps of WRN-16-1 (b) and WRN-40-1 (c). The deep model focuses on more critical region than the shallow one due to its powerful ability.   }\label{fig:attentionmap}
\end{figure}

Our motivation is illustrated in Figure \ref{fig:structure} which trains the student network using two teachers, i.e., expert teacher (black ball) and scratch teacher (red ball). Note that, the three balls start training from the same point due to the same seed. The only difference is that the black ball which represents the expert teacher reaches the local optimum along the red curve in advance. Then we begin to train the student network under the two teachers' guidance and the green curve describes its optimization path. Let us take one point from the student's optimization path to explain. The green ball has been pulled by two forces from the scratch teacher (in red arrow) and expert teacher (in black arrow) respectively. The scratch teacher with strong ability could pull the student towards its path. And the expert teacher pulls the student to focus on the critical region to achieve the final targets. Due to the scratch teacher penalizing the student step by step, the student network goes along the path close to the scratch teacher. As shown in Figure \ref{fig:attentionmap}, though the different structure of student and teacher network, they focus on the approximate region to classify the dog. But the deep teacher network focuses more on critical region (the whole head of dog) for the task than the shallow model. Thus we use the attention mechanism from expert teacher to provides the key hints which could avoid detours. In such manner, the student gets high performance close to the teachers.

As we can see from Figure \ref{fig:optimization}, we prepare the expert teacher using the normal training process in advance which has been described in the blue rectangular. Then we start to feed data (image batch) to our network and the $X_{t-1}$, $X_{t}$, $X_{t+1}$ means three consecutive moments in our training process. The scratch teacher and student use the standard cross entropy loss between softmax outputs and ground truth label respectively. Furthermore, the scratch teacher penalizes the student using $L2$ loss between its temporary logits and the student's logits at every iteration. Note that, only the student's parameters have been updated during the back-propagation of $L2$ loss term. Because the scratch teacher doesn't need to mimic the outputs of student. However, it is difficult to train a deeper student network using knowledge distillation without introducing the intermediate constraint. So we let the expert teacher provide intermediate constraint using the attention loss. It could constrain the student to focus on the critical region where the expert teacher concentrates on. To train the student network, we optimize the total loss function in Eq. \ref{con:inventoryflow}. We will detail the objective function in next section. 

\subsection{Formulation}
Deep neural networks can generate features from any layers. The knowledge distillation technology usually uses different layer's features or outputs as knowledge to transfer from teacher network to student network. The high layer features are mostly closer to the object parts for performing a specific task. However, the lower layer features are usually the typical generic features (i.e., edges and corners). Therefore, we can take the features generated from the lower parts of the DNNs as the intermediate hints. All these features contain valuable dark knowledge which can be transferred to guide student network's training process.

Let us respectively denote $x$ and $y$ as the input of the DNNs and one-hot labels of our architecture. We let $P_{T}$ be the teacher network's softmax output as
$P_{T} = softmax(a_{{T}})$. Specifically, $P_{T}$ is obtained by applying softmax function on the un-normalized log probability values $a_{{T}}$.  Similarly, the same image fed to the student network to get the predictions $P_{S} = softmax(a_{{S}})$. In the intermediate layers of the DNN, we denote the activation tensor $A\in R^{C\times X \times W}$with its corresponding layer. The pairs of teacher and student attention maps are denoted as $F(A_{T}^{j})$ and $ {F(A_{S}^{j})} $  in vectorized form respectively \cite{Zagoruyko2016Paying}. And the standard cross entropy is denoted as $\mathcal {H}$. Hinton et al.~\cite{Hinton2015Distilling} extend previous works by training a compact student network to mimic the output probability distribution of teacher network. They name this informative and representative knowledge as dark knowledge. It contains the relative probabilities of 'incorrect' classification results provided by teacher networks. When we perform knowledge distillation with a temperature parameter $\tau$ the student network will be trained to optimize the following loss function:

\begin{equation}
\mathcal{L}_{KD}(W_{s}) = \mathcal{H}(y_{true},P_{S}) + \lambda\mathcal{H}(P_{T}^{\tau},P_{S}^{\tau})
\end{equation}

Mishra et al. \cite{Mishra2017Apprentice} propose a new perspective view to jointly train a teacher network (full-precision) and a student network (low-precision) from scratch using knowledge distillation. The total loss function is as following:

\begin{equation}
\mathcal{L}(x; W_{t},W_{s}) = \alpha\mathcal{H}(y_{true},P_{T}) + \beta\mathcal{H}(y_{true},P_{S}) + \gamma\mathcal{H}(a_{{T}},P_{S})
\end{equation}

In this case, the teacher and student network both train from scratch. Moreover, the teacher network would continuously guide the student network not only with the final trained logits \cite{Mishra2017Apprentice}. A similar idea has been studied in~\cite{zhou2017Rocket} where the student network and the teacher network share lower layers and training simultaneously. However, the teacher trained from scratch may provide incorrect guidance to student network in the beginning of the training stage. Another fact is that it is difficult to train a deeper student using knowledge distillation without introducing the intermediate constraint. 

To this end, we propose a new knowledge distillation method using two teachers. We denote the expert teacher trained in advance as $T1$ and the scratch teacher with random initialization as $T2$. The $T1$ provides intermediate constraint using attention maps~\cite{Zagoruyko2016Paying} from lower layers using the loss function as following:

\begin{eqnarray}
\mathcal{L}(x; W_{T1},W_{s}) = \sum_{j = 1}^{N_{L}}\parallel \frac{F(A_{S}^{j})}{\parallel F(A_{S}^{j}) \parallel_{2}} - \frac{F(A_{T1}^{j})}{\parallel F(A_{T1}^{j}) \parallel_{2}} \parallel_{2}
\end{eqnarray}

\noindent The $F$ means the activation-based mapping function which inputs the above 3D tensor $A$ and outputs a spatial attention map, i.e., a flattened 2D tensor. More specifically, $F_{sum}^{p}(A) = \sum_{i = 1}^{C} \mid A_{i} \mid ^{p} $ , sum of absolute values raised to the power of $p$ (where $p > 1$). And the $T2$ provides the log probability values before softmax as constraint from every step, i.e., $\lambda\parallel a_{{S}} - a_{{T2}}\parallel_{2}^{2}$. It's important to note that this constraint only affects the back propagation of student network to avoid teacher network closing to student network. When we train the compact student, we aim to optimize the following loss function:

\begin{eqnarray}
\mathcal{L}(W_{s}, W_{T1}, W_{T2} ) = \mathcal{H}(y_{true},P_{S}) + \mathcal{H}(y_{true},P_{T2}) +\\\nonumber
\lambda\parallel a_{{S}} - a_{{T2}}\parallel_{2}^{2} + \beta\sum_{j = 1}^{N_{L}}\parallel \frac{F(A_{S}^{j})}{\parallel F(A_{S}^{j}) \parallel_{2}} - \frac{F(A_{T1}^{j})}{\parallel F(A_{T1}^{j}) \parallel_{2}} \parallel_{2}\label{con:inventoryflow}
\end{eqnarray}

The first part of total loss ensures $T$ and $S$ to train as original manner independently. In the second part, we denote the knowledge distillation loss \cite{Caruana2013Do} as $L2$ loss between logits $a_{{S}}$ and $a_{{T}}$. To optimize the above loss function, the log probability values $a_{{S}}$ from the student network is to mimic the softmax activation $a_{{T}}$ from the teacher network. So the student network benefits from the supervisory information of the teacher network during all the training process. The complex teacher model with more learning capability can provide the possible path towards the final target. The last part from our architecture provides intermediate-level hints from a pre-trained teacher network. 

\begin{algorithm}[t]
\caption{Training with Collaborative Teaching } 
\hspace*{-0.08cm} {\bf Input:} 
image data and label data $(x,y)$.\\
\hspace*{-0.08cm} {\bf Output:} 
parameters $W_{s}$ of student model.\\
\hspace*{-0.08cm} {\bf Initialize:} 
$W_{s}$, $W_{T2}$ and training hyper-parameters.\\
\hspace*{-0.08cm} {\bf Stage 1:} 
Prepare the expert teacher.
\begin{algorithmic}[1]
\State \bfseries Repeat: \mdseries
    \State compute $\mathcal{H}(y_{true},P_{T1})$.
    \State update $ W_{T1} $ by gradient back-propagation.
\State \bfseries Until: \mdseries $\mathcal{H}(y_{true},P_{T1})$ converges.
\end{algorithmic}
\hspace*{-0.08cm} {\bf Stage 2:} 
Training the student collaboratively.
\begin{algorithmic}[1]
\State \bfseries Repeat: \mdseries
    \State compute $\mathcal{L}(W_{s}, W_{T1}, W_{T2} )$ by Eq. \ref{con:inventoryflow}.
    \State update $W_{s}, W_{T2} $ by gradient back-propagation.
\State \bfseries Until: \mdseries $\mathcal{L}(W_{s}, W_{T1}, W_{T2} )$ converges.
\State \Return $ W_{s} $
\end{algorithmic}
\end{algorithm}

\subsection{Training procedure}
The learning procedure contains two stages of training. On the first stage, we minimize the cross entropy loss $ \mathcal{H}(y_{true},P_{T1}) $ to initialize the parameters of expert teacher ($T1$). Then we train the student network using two teachers $T1$ and $T2$ simultaneously by optimizing the total loss function as shown in Eq. \ref{con:inventoryflow}. The learning procedure is explained in Algorithm 1. 

Our proposed method jointly trains the student network using two teachers. It is crucial to combine the temporary outputs from scratch teacher with the intermediate features from the expert teacher in the whole training process. The scratch teacher $T2$ guides the student step by step using the log probability values before softmax. Due to the powerful learning capability of scratch teacher, it makes the student close to the final target following the optimal path. However, only the supervised information from one single scratch teacher is not enough. Because the scratch teacher attempts many paths to find the optimal one. Meanwhile, the student follows it and pace backwards and forwards. Thus we need the expert teacher to provide intermediate hints such as attention maps. With the constraint imposed to intermediate layer, the student can find the correct path not only quick but also definitely. 

We will demonstrate that the student from our knowledge distillation method gets improved performance in Section \ref{sec4}. However, one might ask how the scratch teacher affect the training process of student network? If the scratch teacher works, why not only use it to train student network? Or would other knowledge from the expert teacher be better helpful than attention knowledge? We attempt to investigate these questions from both empirical and theoretical aspects in Section \ref{sec4}.

\section{Experiments}\label{sec4}
In this section, we verify the effectiveness of our proposed CTKD method and investigate the importance of Collaborative Teaching. Experiments are conducted on several standard datasets CIFAR-10, CIFAR-100, SVHN and Tiny ImageNet. We compare our proposed CTKD method with the existing knowledge distillation methods, including knowledge distillation (KD) \cite{Hinton2015Distilling}, Attention Transfer Knowledge Distillation (ATKD) \cite{Zagoruyko2016Paying} and Rocket Launching Knowledge Distillation (RLKD) \cite{zhou2017Rocket}. We implement the networks with Pytorch and trains on 1080Ti GPUs. Note that, there are several hyperparameters in our experiments that need to be consistent. For the original KD method, we set the temperature factor for softened softmax to 4 as in \cite{Hinton2015Distilling}. And the $ \beta $ of AT is set to $10^{3}$ following \cite{Zagoruyko2016Paying}. Code is available at \href{https://github.com/ouc-ocean-group/CTKD}{https://github.com/ouc-ocean-group/CTKD}.

\subsection{Experimental Setup}
\begin{figure}
  \centering
  \includegraphics[width=8.5cm]{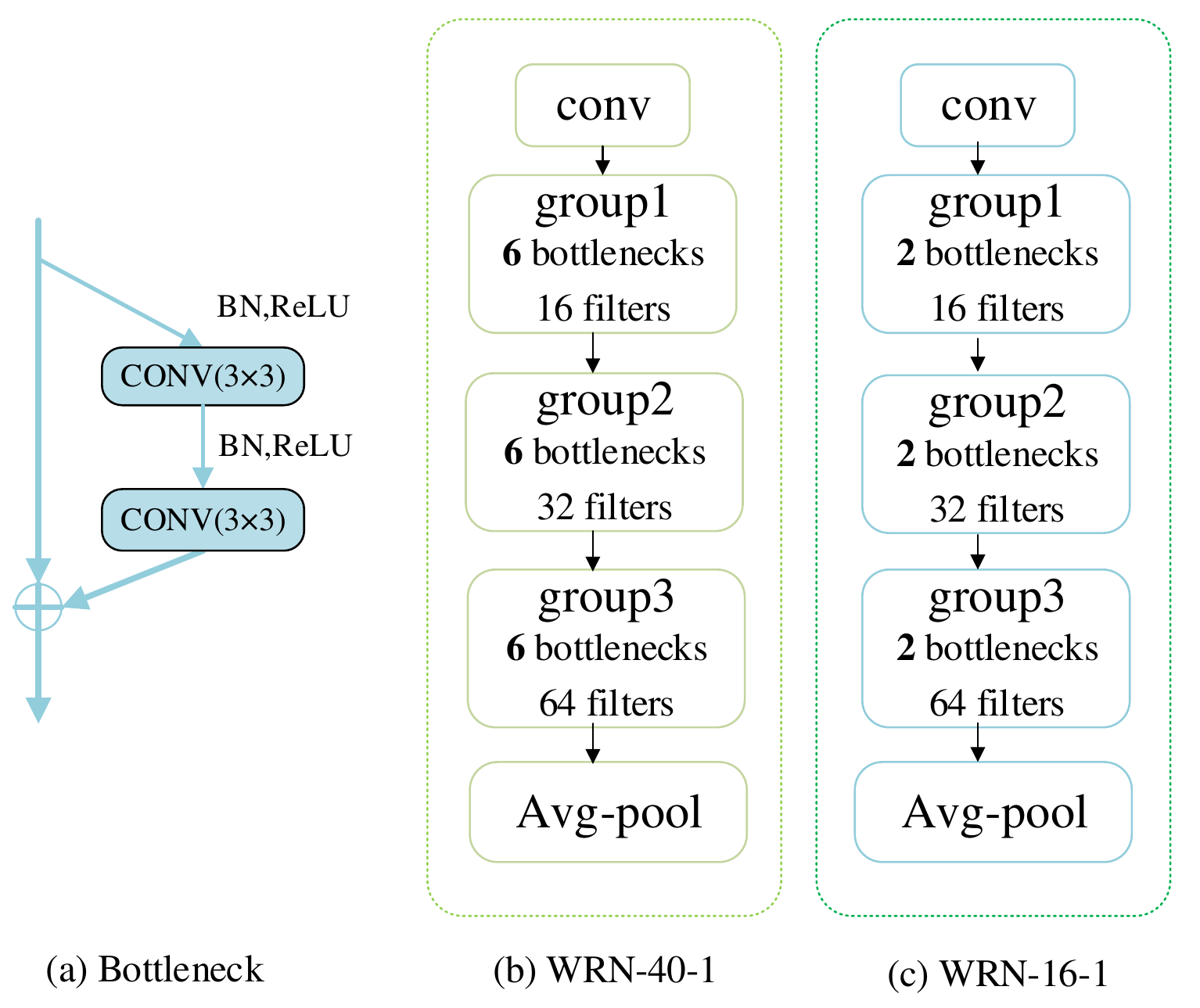}\\
  \caption{Structure of wide residual networks. (a) describe the basic residual blocks which is used in our base architecture. The widen factor m determine the network's width and n means the number of bottlenecks in each group. (b)(c) show a pair of teacher-student network,  WRN-40-1 and WRN-16-1.}\label{fig:wrn}
\end{figure}

\textbf{Network architecture.} For all experiments, we employ the Wide Residual Network (WRN) \cite{Zagoruyko2016Wide} as our base architecture for teacher and student network. The WRN stacks the basic residual blocks \cite{He_2016_CVPR} as shown in Figure \ref{fig:wrn} (a) to achieve state-of-the-art performance. Moreover, it uses the additional widen factor $m$ to increase the width, which could bring more representation ability. The wide residual network has a standard convolutional layer (conv) followed by three groups of residual blocks, each of size n. Furthermore, the total depth and widen factor are served as a proxy for the size or flexibility of the network architecture. In the following sections, the architecture of Wide Residual Networks (WRN) is denoted as WRN-d-m \cite{DBLP:journals/corr/abs-1709-00513}, where the total depth is $d = 6n + 4$, $n$ represents the number of residual blocks and $m$ is the widen factor used to increase the number of filters in each residual block. Our teacher network is deep and wide WRN with large $d$ and $m$, while student network is shallow and thin WRN with small $d$ and $m$. As shown in Figure \ref{fig:wrn} (b)(c), WRN-40-1 is our teacher network and student network uses the WRN-16-1. 

\textbf{Implementation Details.} We firstly conduct our experiments on the public datasets CIFAR-10 which has $32\times32$ small RGB images. For all experiments, we use minibatches of size 128 for training. Moreover, we use horizontal flips and random crops for data augmentations before each minibatch. The learning rate starts with 0.1 and is reduced by a factor of 0.2 on epoch 60, 120 and 160 respectively. For CIFAR dataset, we use stochastic gradient descent with momentum fixed at 0.9 for 200 epochs. However, we use Adam \cite{DBLP:journals/corr/KingmaB14} with learning rate 0.01 initially and drop the learning rate by 0.2 at epoch 20, 40, 60 for SVHN dataset which is easy to learn. Furthermore, all networks have batch normalization.

\begin{figure}
  \centering
  \includegraphics[width=8.5cm]{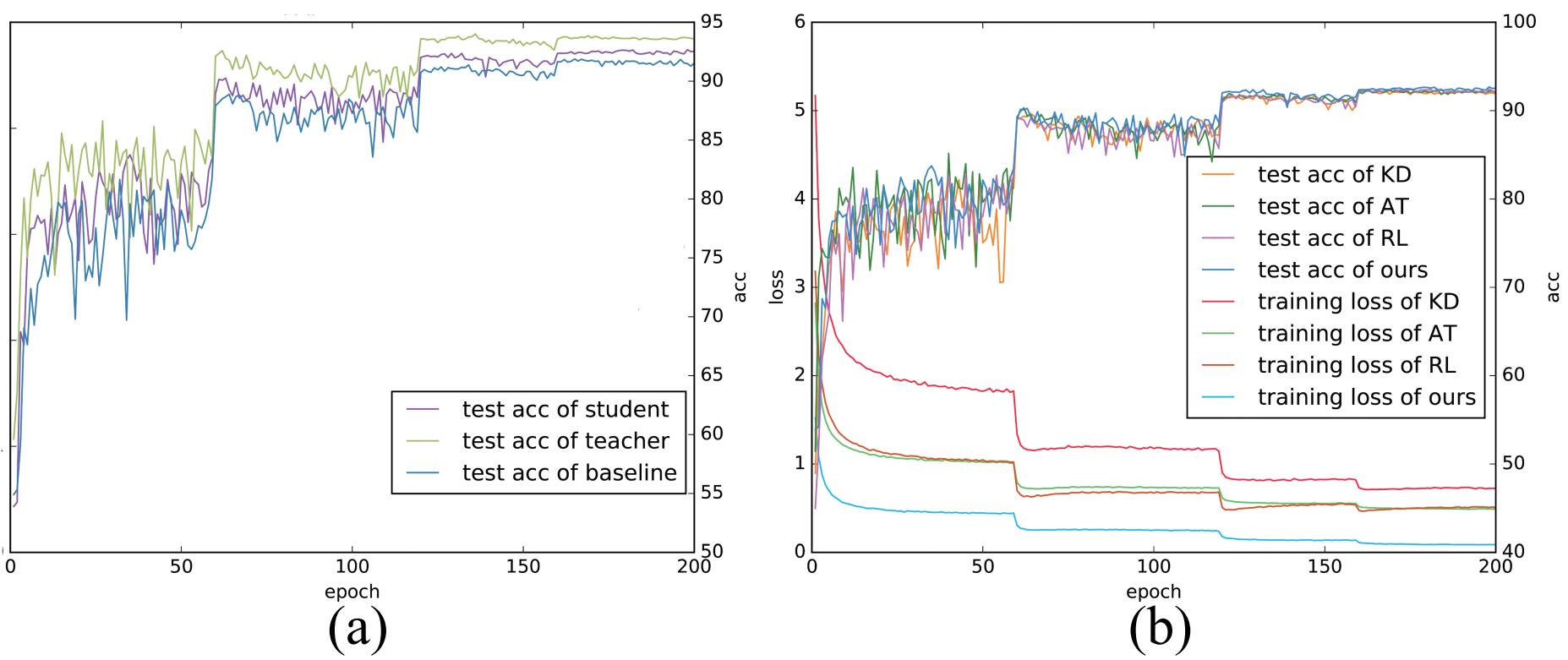}\\
  \caption{(a) the testing accuracy of scratch teacher, student from our knowledge distillation method and student trains individually. (b) Training loss and testing accuracy of different knowledge transfer methods on CIFAR-10.}\label{fig:loss}
\end{figure}

\subsection{CIFAR-10}\label{sec4.2}
The CIFAR-10 dataset \cite{krizhevsky2009learning} contains $32\times32$ small RGB images with 10 classes. It consists of 50K training  images with 5K images per class and 10K testing images with 1K images per class respectively. However, we use the $32\times32$ RGB images after random crops and horizontal flips for training. And the original $32\times32$ RGB images are used for testing.

We use the deep and wide WRN (e.g. WRN-40-1 and WRN-40-2) as the teachers network. However, the student network uses the shallow and thin WRN (e.g. WRN-16-1, WRN-16-2). Note that, we firstly train the expert teacher network using the normal training procedure on CIFAR-10 dataset, which provides 93.43\% accuracy for the classification task. The scratch teacher network and student network are random initialized. We use the scratch and expert teacher network to collaboratively supervise the training of student network as described in Figure\ref{fig:structure}.

\begin{table*}
\begin{center}
\caption{Classification accuracy (\%) on CIFAR-10. Acc is computed as median of 5 runs with different seed. We conduct two groups of experiments, i.e. one (left part) is a Student (WRN-16-1) with Personal Teacher (WRN-40-1) and Teacher (WRN-40-1), one (right part) is a Student (WRN-16-2) with Personal Teacher (WRN-40-2) and Teacher (WRN-40-2). Baseline means the WRN-16 trains individually. CTKD means the WRN-16 results in our CTKD method. Teacher means WRN-40 trains individually in advance. Personal Teacher means WRN-40 trains from scratch with student. }
\label{table:1}      
\begin{tabular}{|c|c|c|c||c|c|c|}
\hline
Type & Model & Params(M) &  Acc(\%) & Model & Params(M) & Acc(\%) \\
\hline
\hline
Baseline & WRN-16-1 & 0.17 & 91.28 &WRN-16-2 &0.69 &93.68\\
KD & WRN-16-1 & 0.17 & 91.60  &WRN-16-2 &0.69 &93.93\\
ATKD & WRN-16-1 & 0.17 &91.77  &WRN-16-2  &0.69&94.11\\
RLKD & WRN-16-1 & 0.17 &91.96  &WRN-16-2  &0.69&94.23 \\
CTKD & WRN-16-1 & 0.17 & \bfseries 92.50 \mdseries &WRN-16-2  &0.69&\bfseries94.42\mdseries\\
\hline
\hline
Scratch Teacher & WRN-40-1 & 0.56 &93.43  &WRN-40-2  &2.20&94.70\\
\hline
\end{tabular}
\end{center}
\end{table*}

\begin{table}
\begin{center}
\caption{Classification accuracy (\%) on CIFAR-10 (5 runs). The student (WRN-16-1) the results with different combinations of teachers in our knowledge distillation architecture. Baseline means the WRN-16-1 trains individually.}
\label{table:2}
\begin{tabular}{|c|c|c|c|c|}
\hline
Model & \tabincell{c}{Scratch \\Teacher} & \tabincell{c}{Expert\\with AT} & \tabincell{c}{Expert \\with KD} & Acc (\%)\\
\hline
\hline
Baseline & -- &-- & -- &91.28\%      \\
KD &--  &-- & \checkmark &91.60\%     \\
ATKD &--  & \checkmark &-- &91.77\%   \\
RLKD & \checkmark &-- & --&91.96\%   \\
RLKD+KD & \checkmark &-- & \checkmark &92.30\%   \\
CTKD & \checkmark & \checkmark &-- & \bfseries 92.50\% \mdseries   \\
\hline
\end{tabular}
\end{center}
\end{table}

From the experimental results in Table ~\ref{table:1}, we can find our proposed Collaborative Teaching Knowledge Distillation (CTKD) method improves the generalization ability of student network and gets notable improvement compared to the existing methods. Note that, all the numbers are the results
of our implementation. We implement KD and ATKD according to \cite{Zagoruyko2016Paying}. We repeat 5 times with different seed and take the median of classification accuracy as the final results for all experiments. We set two pairs of teacher-student, i.e., WRN-16-1 with WRN-40-1 teacher and WRN-16-2 with WRN-40-2 teacher. Taking the left part of the table as an example, we use WRN-16-1 as student network and WRN-40-1 is used as scratch and expert teacher network. We train the expert teacher using normal training procedure independently. And it gets 93.43\% accuracy. Furthermore, the student network using the normal training method shows a 91.28\% recognition rate. Surprisingly, our new architecture of Collaborative Teaching Knowledge Distillation (CTKD) gets 92.50\% accuracy with 1.22\% improvement than the independent student. And the performance of student in our method is close to the teacher network. Moreover, we compare the performance of the student network with existing knowledge distillation method (i.e., KD, ATKD, RLKD). And the proposed method with distilled knowledge clearly performs better than the existing ones. As shown in Figure \ref{fig:loss} (a), the student from our knowledge distillation method gets significant improvement than it trains individually (baseline). And we plot the testing accuracy and training loss curves of all the experiments in Figure \ref{fig:loss} (b). It describes the recognition results of different knowledge transfer methods compared with ours on CIFAR-10 dataset. We can observe that our CTKD method gets a significant improvement on final accuracy and outperforms existing methods. It can be also noticed that our method has a fast convergence speed. This will be further discussed in the next part with more comparisons.

\begin{table}
\begin{center}
\caption{Classification accuracy(\%) on CIFAR-10 (5 runs) with different forms of intermediate knowledge. CTKD \ddag means the WRN-16-1 results initialized through transferring weights from WRN-40-1 for lower layers.}
\label{table:3} 
\begin{tabular}{|c|c|c|}
\hline
 \tabincell{c}{Intermediate\\Knowledge} & \tabincell{c}{WRN-16-1\\ with WRN-40-1} & \tabincell{c}{WRN-16-2\\ with WRN-40-2}   \\
\hline
\hline
FitNet & 91.70\% & 93.98\%    \\
CTKD\ddag & 91.89\% & 94.20\%  \\
CTKD & \bfseries 92.50\% \mdseries & \bfseries 94.42\% \mdseries\\
\hline
\end{tabular}
\end{center}
\end{table}

The improvement of our CTKD method is attributed to both the supervised information from the two teachers. We compare the accuracy of student DNN in our knowledge distillation architecture with different combination of teacher DNNs. As shown in Table \ref{table:2}, for the generalization ability of student DNN, the two teachers are equally important and complement each other. 
The recognition rates of student network under the single guidance of scratch teacher is 91.54\%. It also gets 91.77\% accuracy when we only use the expert teacher's attention maps as supervised information in the training process.
Interestingly, the accuracy of student network gets 92.50\% when we collaboratively train it with scratch teacher network and expert teacher network.

\begin{figure}
  \centering
  \includegraphics[width=8.5cm]{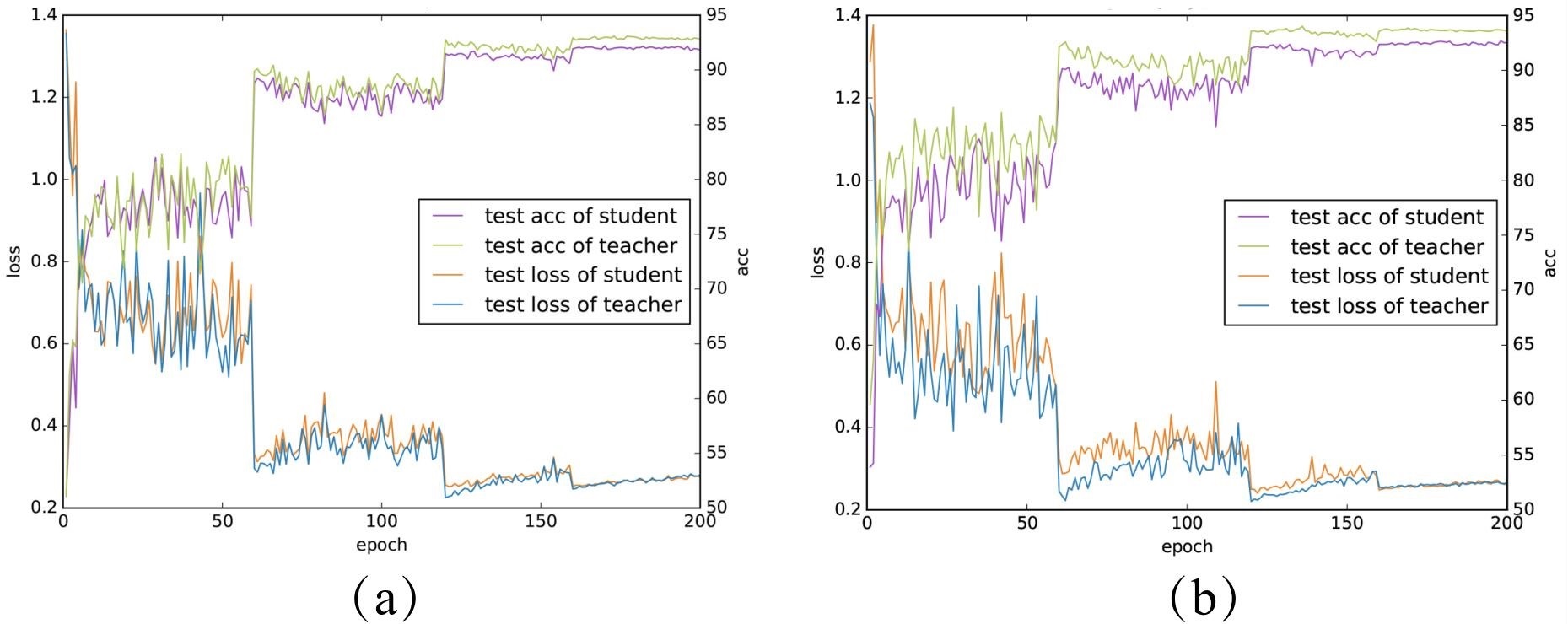}\\
  \caption{(a) testing accuracy and loss of the teacher and student network in \cite{zhou2017Rocket}. (b) testing accuracy and loss of the scratch teacher and student network in our method.}\label{fig:cifar10loss}
\end{figure}

The scratch teacher could provide its temporary outputs of logits to guide the student towards its optimization path. To prove this, we train the student network under the simple guidance of scratch teacher as RLKD \cite{zhou2017Rocket}. As Figure \ref{fig:cifar10loss} (a) shown, the testing accuracy curve of student tightly follows the scratch teacher's. However, we can find that the performance of teacher in \ref{fig:cifar10loss} (a) has been affected due to the parameters sharing on lower layers. The performance of teacher network also limits the student's results. However, our method which introduces the expert teacher improves this in Figure \ref{fig:cifar10loss} (b).

\begin{table*}
\begin{center} 
\caption{Classification accuracy (\%) on SVHN and CIFAR-100 datasets (5 runs). Baseline means the WRN-16 trains individually. CTKD means the WRN-16 results in our method.}
\label{table:4}
\begin{tabular}{|c|c|c|c|c|c|c||c|c|}
\hline
Dataset & Model(S/T) & Baseline &  KD  & ATKD & RLKD & CTKD & Teacher\\
\hline
\hline
SVHN &\tabincell{c}{WRN-16-1(0.17M)\\ WRN-40-1(0.56M)} & 94.48 & 94.59 & 94.91&95.77 &\bfseries95.83\mdseries &95.89  \\
\hline
\hline
CIFAR-100 &\tabincell{c}{WRN-16-2(0.69M)\\ WRN-40-2(2.20M)} & 72.27 & 72.54 &72.98 &73.20 &\bfseries74.70\mdseries & 75.42 \\
\hline
\end{tabular}
\end{center}
\end{table*}

Why we use the attention maps as intermediate knowledge from the expert teacher network? We expect that the student could focus on the key region as same as the expert teacher model in the whole training process. As shown in Figure \ref{fig:attentionmap}, we visualize the top-level activation attention maps of pre-trained WRN-40-1 and WRN-16-1 on ImageNet dataset using the visualization technique in \cite{DBLP:journals/corr/SelvarajuDVCPB16}. We can observe that the attention maps from different depth models focus on different region. Specifically, the deeper teacher model with powerful ability focuses on the pivotal region in order to classify the input image, however the shallow student model focuses on a wider area. Thus we make the student network to mimic the attention maps from the expert teacher network. In the training process, the student network learns to focus on the key region under the guidance of expert teacher using the attention maps.

\begin{figure}
  \centering
  \includegraphics[width=8.5cm]{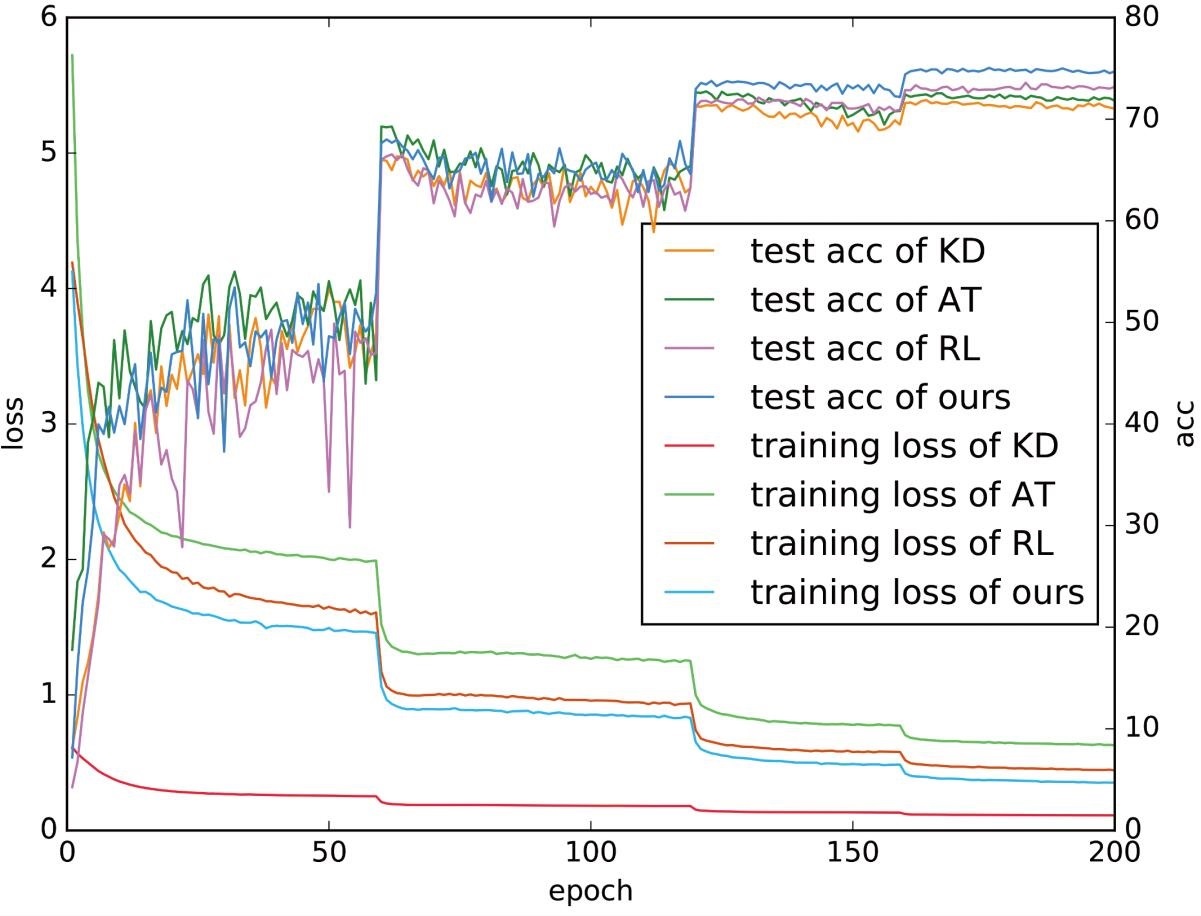}\\
  \caption{ The training loss and testing accuracy of different knowledge distillation approaches on CIFAR-100 dataset.}\label{fig:training loss}
\end{figure}

To demonstrate the effectiveness of attention mechanism in our collaborative teaching architecture, we transfer different forms of intermediate knowledge from the expert teacher to student network. Fitnet  \cite{Romero2015FitNets} provides a kind of intermediate supervised knowledge, i.e., the features maps from the middle layers of DNNs. Another form of intermediate knowledge can be the weights transferred from the teacher network for the student network, due to the same architecture in both teacher and student model. But these supervised information may be a hard constraint for the student network. Table \ref{table:3} shows the accuracy of student network using different intermediate knowledge transferred from the middle outputs of the expert teacher network in our architecture. The first row means the accuracy of student network when expert teacher using the intermediate knowledge in FitNet \cite{Romero2015FitNets}. As shown, its performance is slightly better than the individual one's. And the second row shows the student network which directly transferring weights from the expert teacher for initializing also gets slightly improvements. The student network from our proposed knowledge distillation method in the last row gets best performance. Because the attention maps just hint the student network to focus on the key region instead of imposing hard constraint for the lower layers.

\subsection{CIFAR-100 and SVHN}\label{sec4.3}
In this section, we verify the effectiveness of our proposed method through conducting classification task on CIFAR-100 and SVHN dataset. 

The CIAR-100 dataset \cite{krizhevsky2009learning} contains 50K training images and 10K testing images. However, it contains 100 classes which is more challenge than CIFAR-10. Due to more complicated classification tasks, we set the width factor to 2 for our WRN architecture. Thus we use WRN-40-2 as the teacher network and WRN-16-2 is used as the student network. 

The SVHN dataset \cite{netzer2011reading} is similar to MNIST with small $32\times32$ RGB cropped digits in 10 class and it is obtained from house numbers in Google Street View images. SVHN has 73257 images for training, 26032 images in testing set and 531131 samples additional. 

As shown in Table \ref{table:3}, the student network (WRN-16-2) from our CTKD method achieves 74.70\% classification accuracy on CIFAR-100 dataset and gets 2.43\% improvement compared with the student network trained individually. We also compare our proposed CTKD method with some of the most recent state-of-the-art knowledge distillation methods. We can see that the student collaboratively trained from our proposed method outperforms all of them. Figure \ref{fig:training loss} shows the accuracy change curves over time among different knowledge distillation methods on CIFAR-100. Interestingly, we observe that our method has a significant improvement than used on CIFAR-10 dataset through comparing the Figure \ref{fig:training loss} and Figure \ref{fig:loss} (b). Considering that the CIFAR-100 dataset and WRN(wide factor as 2) is more complicated than CIFAR-10, we believe that our method is an effective technique for transferring the knowledge to compact network.  We use the Adam with learning rate 0.01 initially for SVHN dataset as implementation details described and train the network 100 epochs. Furthermore, the student (WRN-16-1) also achieves 1.35\% improvement compared with the baseline.

\subsection{Tiny ImageNet}\label{sec4.4}
We also validate the proposed method through conducting image classification task on a much more challenging dataset, Tiny ImageNet dataset~\cite{le2015tiny}, which is a popular subset of the ImageNet database~\cite{russakovsky2015imagenet}. Tiny Imagenet contains $64\times64$ sized images with 200 classes. Each class has 500 training images, 50 validation images, and 50 test images. 

In our Tiny ImageNet classification experiments, we apply random rotation and horizontal flipping for data augmentation. We optimize the model using stochastic gradient descent(SGD) with mini-batch 128 and momentum 0.9. The learning rate starts from 0.1 and is multiplied by 0.2 at 60, 120, 160, 200, 250 epochs. We totally train the network for 300 epochs and adopt the deep and wide WRN (WRN-40-1) for a teacher model and WRN-16-1 as a student model.

Table \ref{table:5} shows the classification results on Tiny ImageNet. The student network (WRN-16-1) from our CTKD method achieves 53.59\% classification accuracy and gets 2.94\% improvement compared with the student network trained individually. The overall results show that the proposed CTKD method outperforms the recent state-of-the-art knowledge distillation methods.  

\begin{table}
\begin{center}
\caption{Classification accuracy (\%) on Tiny ImageNet (5 runs). Baseline means the WRN-16-1 trains individually. CTKD means the WRN-16-1 results in our method.}
\label{table:5}
\begin{tabular}{|c|c|c|c|c|}
\hline
Type & Model & Params(M) & Acc (\%)\\
\hline
\hline
Baseline & WRN-16-1 & 0.17 & 50.65  \\
KD &  WRN-16-1 & 0.17  &  51.26\\
ATKD &  WRN-16-1 & 0.17 &  52.11\\
RLKD &  WRN-16-1 & 0.17 &  52.54\\
CTKD &  WRN-16-1 & 0.17 & \bfseries53.59\mdseries \\
\hline
\hline
Teacher &  WRN-40-1 & 0.56 & 56.51 \\
\hline
\end{tabular}
\end{center}
\end{table}

\subsection{Analysis of the proposed method}
Most existing well-performed knowledge distillation methods force the compact student to mimic the pre-trained teacher's outputs. However there is a gap between the shallow student network and the deep teacher network due to their different network structure. It could be a hard constraint to learn the pre-trained teacher's knowledge for the student network. Thus we use a scratch teacher to supervise the training of student using every step's temporary outputs. The scratch teacher provides optimal path information to the student network as in Figure \ref{fig:cifar10loss} (a). Moreover, the expert teacher only provides the key hints using attention maps for lower layers which close to the common features. This indicates that the student network will be trained under collaboratively supervising from two teachers. As shown in Figure \ref{fig:cifar10loss} (b), the student and teacher network both get a higher performance than the method~\cite{zhou2017Rocket} as shown in Figure \ref{fig:cifar10loss} (a). 

Why does our collaborative teaching approach work? Firstly, the scratch teacher could transfer its path information to the student on every step as shown in Figure. \ref{fig:cifar10loss} (a). Though it could make mistakes in its training process, at least it provides a path to higher performance than student. Secondly, the expert teacher could also provide additional supervising information to the student network. However, which kind of knowledge from the expert teacher is most effective and suitable in our collaborative teaching approach? We investigate the effects of different knowledge which the expert teacher provides in our structure. The attention mechanism achieves excellent results. The expert teacher only provides the information about where it looks to the student network in the training process. Despite the student's weaker ability, the expert teacher's information makes it possible to catch up with the scratch teacher. We verify the effectiveness of our method with most existing knowledge distillation approaches on CIFAR-10, CIFAR-100, SVHN and Tiny ImageNet datasets in section \ref{sec4.2}, \ref{sec4.3}, \ref{sec4.4}.    

\section{Conclusion}

In this paper, we propose a novel and efficient knowledge distillation method to train a compact student neural network, which can be directly deployed on the resource-constrained devices. We show that the scratch teacher and expert teacher could provide different knowledge from training process and results. To fully utilize both of these knowledge, we propose the Collaborative Teaching Knowledge Distillation (CTKD) method for transferring knowledge from teachers to student network. In detail, we use the scratch teacher to supervise every step of the student's training process. It can guide the student towards the final logits with high accuracy step by step along the optimization path. And the expert teacher only constrains the student to focus on the critical region in the whole training process. In such manner, the compact student network can produce performance closely to the teacher. We  compare  our  proposed  CTKD  method  with  the  state-of-the-art knowledge  distillation  methods. Experimental results show that our method has a significant improvement for student network's classification recognition on CIFAR-10, CIFAR-100, SVHN and Tiny ImageNet datasets. We believe our method is a
valuable complement to the state-of-the-art.

\section*{Acknowledgment}
We thank supports of National Natural Science Foundation of China under Project No. U1706218 and 41576011.

\bibliographystyle{IEEEtran}
\bibliography{IEEEexample}

\end{document}